%% file: main.tex
\documentclass{article}

\usepackage[preprint]{neurips_2026}

\input{macros}

\title{\method{}: Goal-Conditioned Model Predictive Control with Object-Centric Representations}

\author{%
  Jonathan~Spieler\thanks{Correspondence to \texttt{spieler@ais.uni-bonn.de}.} \quad\quad
  Angel~Villar-Corrales \quad\quad
  Sven~Behnke\\
  \parbox[t]{0.9\textwidth}{\centering\small{Autonomous Intelligent Systems, Computer Science Institute VI - Intelligent Systems and Robotics, Center for Robotics and the Lamarr Institute for Machine Learning and Artificial Intelligence, University of Bonn, Germany}}
}

\begin{document}

\maketitle
\vspace{-1em}

\begin{abstract}
	Predictive world models enable agents to model scene dynamics and reason about the consequences of their actions.
	Inspired by human perception, object-centric world models capture scene dynamics using object-level representations, which can be used for downstream applications such as action planning.
	However, most object-centric world models and reinforcement learning (RL) approaches learn reactive policies that are fixed at inference time, limiting generalization to novel situations.
	We propose \method, an object-centric world modeling framework that
	enables planning through Model Predictive Control (MPC).
	\method leverages vision encoders to learn slot-based representations, which encode individual objects in the scene, and uses these structured representations to learn an action-conditioned object-centric dynamics model.
	At inference time, the learned dynamics model enables action planning via MPC, allowing agents to adapt to previously unseen situations.
	Since the learned world model is differentiable, we can use gradient-based MPC to directly optimize actions, which is computationally more efficient than relying on gradient-free, sampling-based MPC methods.
	Experiments on simulated robotic manipulation tasks show that \method
	improves both task performance and planning efficiency compared to non-object-centric world model baselines.
	In the considered offline setting with limited state-action coverage, we find that gradient-based MPC performs better than gradient-free, sampling-based MPC.
	Our results demonstrate that explicitly structured, object-centric representations provide a strong inductive bias for controllable and generalizable decision-making.
	Code and additional results are available at \projectpage.
\end{abstract}

\section{Introduction}
\label{sec:introduction}
The ability to predict how the world evolves in response to actions is fundamental to intelligent behavior.
World models \citep{ha_world_2018, lecun_path_2022} aim to equip artificial agents
with this capability by learning predictive models that enable forecasting future environment states, supporting anticipation and planning.
Humans, however, do not perceive the world as an unstructured stream of pixels.
Instead, we parse scenes into persistent objects, which move independently, can interact with each other, or compose into more complex entities \citep{Kahneman_ReviewingOfObjectFiles_1992}.
Inspired by human perception, several works have investigated object-centric and compositional inductive biases for predictive modeling, demonstrating desirable properties such as generalization to novel compositions \citep{Haramati_EntityCentric_2024}, transferability to novel tasks~\citep{Zhang_IsAnObject-CentricVideoRepresentationBeneficialForTransfer_2022}, and improvements in sample efficiency~\citep{sold2025mosbach}, among others.

\begin{figure}[ht]
  \setlength{\belowcaptionskip}{5pt}
  \centering
  \begin{subfigure}[ht]{0.425\textwidth}
    \includegraphics[width=\textwidth]{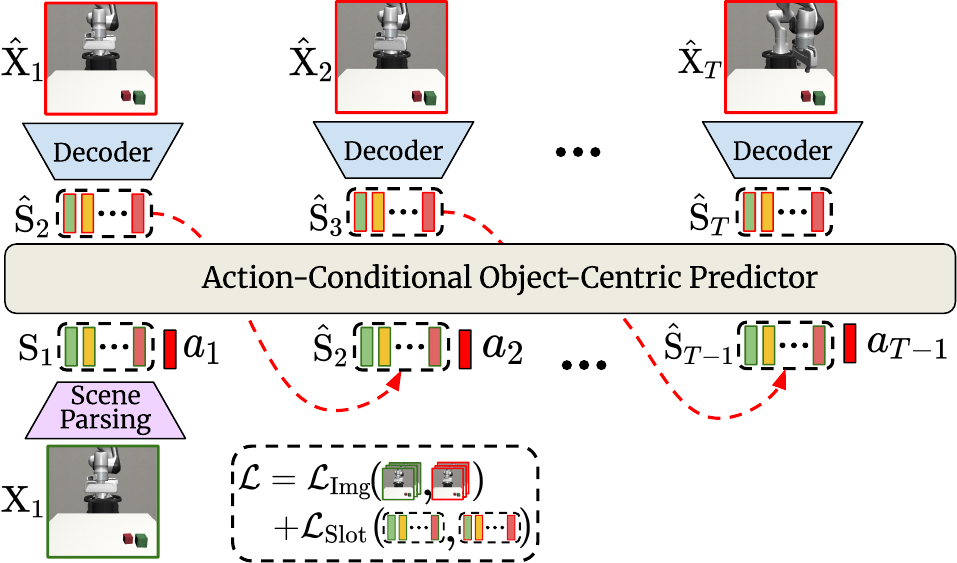}
    \caption{cOCVP training}
    \label{fig:cocvp}
  \end{subfigure}%
  \hfill
  \begin{subfigure}[ht]{0.55\textwidth}
    \includegraphics[width=\textwidth]{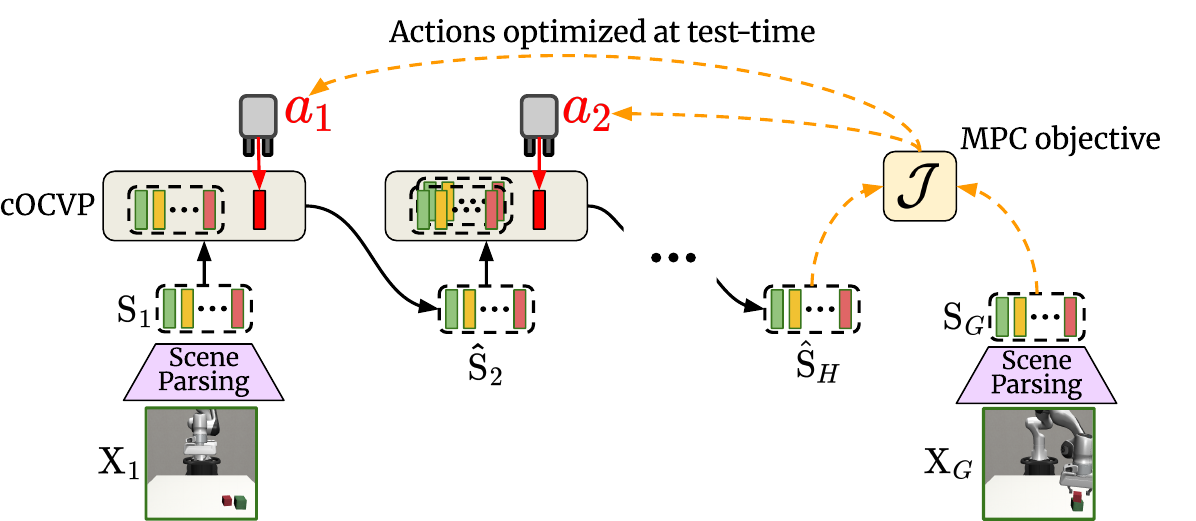}
    \caption{\method inference}
    \label{fig:mpc}
  \end{subfigure}
  \caption{Overview of \method. \textbf{(a)} The object-centric world model (cOCVP) is trained given a single video frame $\ImageT{1}$ and actions $\Action$, and autoregressively predicts future video frames and slot representations $\Slots$. \textbf{(b)} \method{} parses an image $\ImageT{1}$ into its object representations $\SlotsT{1}$ and then predicts the future object states over the horizon $\Horizon$ using cOCVP given actions $\Action$, which are provided by MPC. The goal image $\ImageT{G}$ is also parsed into its object representations $\SlotsT{G}$ and the slots at time step $\PredSlotsT{T}$ and $\SlotsT{G}$ are used to optimize the actions using the MPC objective $\Objective$ defined in \cref{eq:mpc-objective}.}
  \label{fig:teaser}
\end{figure}

Based on these insights, the field of object-centric learning has rapidly progressed in recent years, moving from learning object representations on synthetic images \citep{Burgess_MonetUnsupervisedSceneDecompositionRepresentation_2019,Locatello_ObjectCentricLearningWithSlotAttention_2020} and videos \citep{Kipf_ConditionalObjectCentricLearningFromVideo_2022}, towards more complex real-world scenarios and datasets \citep{Seitzer_BridgingTheGapToRealWorldObjectCentricLearning_2023,Zadaianchuk_VideoSaur_2024}.
Additionally, the learned object representations have been used for world modeling and robotic tasks, including model-based RL~\citep{Ferraro_Focus_2023,sold2025mosbach} or imitation learning~\citep{villar_PlaySlot_2025}.

While effective, most object-centric RL policies are fixed at inference time,
leading to purely reactive behaviors that limit generalization to novel situations~\citep{byravan_evaluating_2021}.
Furthermore, online RL approaches are typically not sample-efficient and require large numbers of
environment interactions, making them costly.
Planning-based control offers a complementary alternative.
Model Predictive Control~\citep{cutler_dynamic_1979,richalet_model_1978} (MPC)
enables online planning by optimizing actions using a learned dynamics model at inference time.
Recent works such as TD-MPC2 \citep{hansen_td-mpc2_2024} combine policy networks with MPC to leverage the benefits of both paradigms.

In this work, we propose \method, a novel object-centric framework for goal-conditioned planning with MPC.
Rather than relying only on a reactive policy, \method learns a structured latent world model that enables online planning directly in an object-centric representation space.
At inference time, as illustrated in \cref{fig:mpc}, \method parses environment observations into a
set of object slots, which represent individual entities present in the scene.
Using an action-conditioned object-centric dynamics model, \method predicts future object states from past slots and candidate action sequences, and optimizes actions via MPC by minimizing the distance between predicted and goal object configurations in slot space.
Leveraging the differentiability of the learned world model for efficient optimization, candidate action sequences are directly optimized by gradient descent.
This formulation enables planning directly in a structured object-centric latent representation, rather than relying on holistic scene features.

In our experiments, we demonstrate that \method learns meaningful object representations, which enable more efficient planning compared to holistic, non-object-centric representations. Slot-based object-centric models reduce the latent space dimensionality by 99\% compared to patch-based approaches such as DINO-WM~\citep{zhou_dino-wm}, which leads to more efficient planning.
Furthermore, the explicit disentanglement of object-level entities enables more direct and controllable reasoning 
over scene dynamics, which is particularly beneficial in low-data and low-compute regimes.

In summary, our contributions are as follows:
\begin{itemize}[leftmargin=16pt]
	\item We introduce \method, a gradient-based MPC method that uses a latent object-centric dynamics model and slot-based representations for goal-conditioned planning from purely visual inputs.
	\item We show that the learned slot-based representations enable more efficient gradient-based planning compared to holistic and patch-based latent representations.
	\item We demonstrate that \method successfully solves complex, long-horizon planning tasks that remain 
	challenging for non-object-centric approaches.
\end{itemize}

\section{Related Work}
\label{sec:related-work}

\paragraph{Slot-Based Object-Centric Learning:}
Slot-based object-centric models aim to decompose visual scenes into a
set of $\NumSlots$ latent embeddings, referred to as \emph{slots},
where each slot represents a distinct object or entity in the
scene~\citep{Locatello_ObjectCentricLearningWithSlotAttention_2020,Greff_OnTheBindingProblemInNeuralNetworks_2020}.
Early approaches learned such object-centric representations
end-to-end from synthetic images~\citep{Locatello_ObjectCentricLearningWithSlotAttention_2020,Burgess_MonetUnsupervisedSceneDecompositionRepresentation_2019,singh2021illiterate}
and videos~\citep{Kipf_ConditionalObjectCentricLearningFromVideo_2022,singh2022simple,zoran2021parts},
demonstrating unsupervised scene decomposition
under controlled settings.
More recent methods leverage pretrained vision encoders~\citep{Seitzer_BridgingTheGapToRealWorldObjectCentricLearning_2023,Zadaianchuk_VideoSaur_2024,aydemir2023self} or additional supervision~\citep{elsayed2022savi++,bao2023object} to enable object-centric learning on complex real-world scenes.
Beyond representation learning, slot-based representations have been
shown to benefit a range of downstream task, including model-based RL~\citep{sold2025mosbach,Haramati_EntityCentric_2024}, visual-question-answering~\citep{Mamaghan_ObjectCentricRepsForVQA_2025} or imitation learning~\citep{villar_PlaySlot_2025}.

\paragraph{Model-based RL and MPC:}

Model-based RL aims to learn a predictive model of environment dynamics, which can then be used to improve decision-making, either by learning policies from imagined experience~\citep{hafner_dream_2020, hafner_mastering_2025}, or by
planning actions at inference time~\citep{sutton_dyna_1991}.
Unlike model-free approaches, which learn reactive policies directly from reward signals, model-based methods explicitly reason about future state evolution, enabling improved sample efficiency and adaptability to new situations.
However, purely policy-based solutions present limited generalization capabilities as their learned policies are solely reactive~\citep{byravan_evaluating_2021}.
Recent works, such as TD-MPC2 \citep{hansen_td-mpc2_2024}, combine policy networks with MPC for online trajectory optimization at inference time, demonstrating strong performance for a broad variety of different environments.
In practice, planning is typically performed using gradient-free,
sampling-based optimization methods such as the Cross-Entropy Method
(CEM)~\citep{rubinstein_optimization_1997} or Model Predictive Path
Integral control (MPPI)~\citep{williams_mppi_2015}.
Recent work~\citep{Sobal2025Learning} shows the potential of gradient-free MPC (MPPI) with a latent dynamics model learned from reward-free offline data on a set of navigation tasks compared to goal-conditioned RL. 

\paragraph{Object-Centric World Models and RL:}
Object-centric world models aim to explicitly model object dynamics and
interactions in video sequences in order to forecast future object and scene
states~\citep{Villar_OCVP_2023,Wu_SlotFormer_2023}.
By representing scenes as compositions of persistent entities, these approaches provide a structured abstraction that is naturally suited for reasoning and decision-making.
Object-centric representations have been explored for downstream
control, both in model-free RL~\citep{zadaianchuk2020self,mambelli2022compositional} and model-based settings~\citep{Ferraro_Focus_2023,sold2025mosbach, Haramati_EntityCentric_2024}.
However, despite progress in reinforcement learning, the integration of slot-based representations with downstream planning remains relatively underexplored.
Early approaches such as O2P2~\citep{janner_reasoning_2019} and
OP3~\citep{veerapaneni_2020} model object interactions but do not
employ modern slot-based representations and rely on predefined
high-level actions rather than low-level control signals.
Moreover, planning in these works is typically performed using
sampling-based, gradient-free optimization methods.
Concurrently to our work, \citet{nam2026causaljepalearningworldmodels} propose a slot-based world model evaluated in an MPC setting on the \PushT{} manipulation task.
While conceptually similar, their approach is tightly coupled to the JEPA~\citep{lecun_path_2022} architecture, does not match the performance of recent holistic world models such as DINO-WM~\citep{zhou_dino-wm}, and also relies solely on sampling-based MPC.

\paragraph{Gradient-based MPC:}
Sampling-based MPC methods typically evaluate hundreds or thousands of candidate trajectories at each decision step, resulting in substantial computational cost.
When dynamics models are represented by differentiable neural networks, it is therefore natural to instead optimize action sequences directly using gradient-based optimization.
The idea of gradient-based planning dates back to the 1960's~\citep{kelley_gradient_1960}, yet
its successful application with learned world models has remained limited.
Prior approaches often rely on large amounts of expert demonstrations~\citep{srinivas_universal_2018}, are only evaluated on low-dimensional domains~\citep{bharadhwaj_model-predictive,s_v_gradient-based_2023},
or struggle to scale to realistic robotics settings~\citep{henaff_model-based_2018}.
Despite their potential benefits, most recent gradient-based MPC methods using learned world
models empirically underperform their sampling-based counterparts, and incur
high computational cost due to a large number of optimization iterations~\citep{zhou_dino-wm,parthasarathy2025closingtraintestgapworld,terver2026drivessuccessphysicalplanning}.
Recent work such as Dream-MPC~\citep{spieler2026dreammpc} revisits this
direction by combining gradient-based planning with learned reward and
value functions within a model-based reinforcement learning framework.
In contrast, we consider an object-centric dynamics model learned from
offline, reward-free data and formulate planning as minimizing the
distance between structured slot representations.
This formulation removes the need for environment interaction during training and allows learning task-agnostic world models that can be reused across tasks within the same environment, enabling trajectory optimization at inference time without retraining.

\section{\method{}}
\label{sec:method}

We propose \method, a novel method that combines gradient-based MPC with a slot-based object-centric latent dynamics model.
\cref{fig:teaser} illustrates the main components of our approach, as well as its training (\cref{fig:cocvp}) and inference (\cref{fig:mpc}).
\method uses a \emph{Scene Parsing} module (\cref{sec:ocvp}) to decompose an image $\ImageT{t}$ into object representations, called slots $\SlotsT{t} = (\SlotSingleT{t}{1}, ..., \SlotSingleT{t}{\NumSlots}) \in \mathbb{R}^{\NumSlots \times \SlotDim}$, where $\NumSlots$ denotes the number of slots and $\SlotDim$ their dimensionality.
Subsequently, a \emph{Conditional Object-Centric Predictor} (cOCVP) autoregressively forecasts future object states over the prediction horizon $\Horizon$, conditioned on the initial parsed object slots and an action sequence, which can be randomly initialized or produced by a learned policy (\cref{sec:policy}).
Given a goal image, the slots predicted at the final rollout step are compared with the goal slots \mbox{-- obtained by parsing the goal image --} in order to optimize the action sequence via the
MPC objective defined in \cref{eq:mpc-objective} (\cref{sec:mpc}).

\subsection{Object-Centric Latent Dynamics Learning}
\label{sec:ocvp}

\method builds upon SAVi~\citep{Kipf_ConditionalObjectCentricLearningFromVideo_2022}, a recursive encoder--decoder model that parses each frame of a video sequence into temporally aligned object representations.
At time step $t$, SAVi encodes the input frame $\ImageT{t}$ into $N$ permutation-equivariant object embeddings $\SlotsT{t} \in \mathbb{R}^{\NumSlots \times \SlotDim}$,
and uses \emph{Slot Attention}~\citep{Locatello_ObjectCentricLearningWithSlotAttention_2020} to iteratively refine the previous slot representations conditioned on image features $\FeatureMaps_t \in \mathbb{R}^{L \times D_{\FeatureMaps{}}}$, where $L$ denotes the number of spatial feature locations and $D_{\FeatureMaps{}}$ their dimensionality.
Specifically, Slot Attention performs cross-attention between slots and image features, with attention coefficients normalized over the slot dimension in order to encourage slots to compete for representing feature locations:
\vspace{-0.1cm}
\begin{align}
	& \Attention = \softmax_{\NumSlots} \left( \frac{q(\SlotsT{t-1})\cdot k(\FeatureMaps_t)^T}{\sqrt{\SlotDim}} \right) \in \R^{\NumSlots \times \NumLocs},
	\label{eq:slot-attention}
\end{align}
\hspace{-0.1cm}
where $k$ and $q$ are learned linear projections.
The slots are then independently updated via a shared Gated Recurrent Unit (GRU)~\citep{Cho_GRU_2014} followed by a residual Multi-Layer Perceptron (MLP):
\begin{align}
	& \SlotsT{t} = \text{GRU}(\Attention \cdot v(\FeatureMaps_t), \SlotsT{t-1})
	\; , \;\; \Attention_{n,l} = \frac{\Attention_{n,l}}{\sum_{i=0}^{\NumLocs-1}\Attention_{n,i}},
	\label{eq:slot-update}
\end{align}
where $v$ is a learned linear projection.
The computations described in \cref{eq:slot-attention,eq:slot-update} can be repeated multiple times with shared weights to iteratively refine the slot representations, producing an accurate object-centric representation of the scene.

To reconstruct images from slots, SAVi employs a slot decoder module.
Namely, each slot is independently processed by a Spatial Broadcast Decoder~\citep{watters2019spatial} ($\SlotDecoder$) to produce an object image and mask, which can be normalized and combined via a weighted sum to synthesize a video frame:
\begin{align}
	& \ObjectImage, \ObjectMask = \SlotDecoder(\SlotSingleT{t}{n}), \;\; \;   \forall \; \SlotSingleT{t}{n}  \;  \in \;  \SlotsT{t},
	\\
	& \PredImageT{t} = \sum_{n=1}^{\NumSlots} \ObjectImage \cdot \ObjectMaskNorm
	\;\;\; \text{with} \;\; \ObjectMaskNorm = \softmax_{\NumSlots}(\ObjectMask).
\end{align}

SAVi is trained self-supervised using an image reconstruction loss:
\begin{align}
	\Loss_{\text{SAVi}} &= 
	\sum_{t=1}^{\NumFrames}|| \SlotDecoder(\ImageEncoder(\ImageT{t})) - \ImageT{t} ||_2^2, \label{eq: loss savi}
\end{align}
where $\ImageEncoder$ and $\SlotDecoder$ correspond to the scene parsing and object rendering modules, respectively.

%
Inspired by OCVP~\citep{Villar_OCVP_2023}, 
we adopt a transformer-based \citep{Vaswani_AttentionIsAllYouNeed_2017} latent dynamics model that autoregressively predicts future objects slots conditioned on past object states.
The model leverages self-attention to capture object dynamics and agent-object interactions while preserving the permutation equivariance of the slot representations.

We extend the OCVP predictor to an action-conditional setting (cOCVP), enabling the model to explicitly account for control inputs when forecasting future object dynamics.
At each time step, action vectors $\ActionT{t} \in \R^{N_a}$ are mapped into the predictor embedding space through a learnable linear projection $\ActionProj$ and combined additively with the slot representations to form the predictor input.
Given past object slots and the corresponding action sequence, the model autoregressively estimates the next slot state:
\begin{align}
	& \PredSlotsT{t+1} = \text{cOCVP}\big(
			\SlotsT{1}, \ActionProj(\ActionT{1}),
			\ldots,
			\SlotsT{t}, \ActionProj(\ActionT{t})
		\big).
\end{align}
Starting from a single seed frame, this process is applied autoregressively by feeding predicted
slots back as inputs, allowing future states to be generated over
a prediction horizon $\Horizon$.

Given the pretrained SAVi model, we train cOCVP by minimizing a combined objective:
\begin{align}
	\Loss_{{\text{cOCVP}}} &=
	\sum_{t=2}^{\NumPreds+1}
	\underbrace{\lambda_{\text{Img}} \cdot || \PredImageT{t} - \ImageT{t} ||_2^2}_{\text{future frame prediction}}
	\; + \;
	\underbrace{\lambda_{\text{Slot}} \cdot || \PredSlotsT{t} - \ImageEncoder(\ImageT{t}) ||_2^2}_{\text{joint-embedding alignment}}
	,
\end{align}
where $\lambda_{\text{Img}} $ and $\lambda_{\text{Slot}}$ are scalar coefficients that balance the contribution of each loss term.
The first loss penalizes frame prediction errors, encouraging accurate visual forecasting, whereas the second term aligns predicted slots with the object-centric representations
inferred from the corresponding ground-truth frames, stabilizing latent dynamics learning and improving temporal consistency.

\subsection{Policy Learning}
\label{sec:policy}

Initializing MPC from randomly sampled action sequences frequently
results in suboptimal solutions, since the optimizer lacks a
meaningful starting point in high-dimensional action spaces.
Prior work has shown that warm-starting MPC with an informed initial
action proposal can substantially improve convergence and optimization
stability~\citep{parmas_pipps_2018,hansen_temporal_2022}.
Motivated by this observation, we learn a policy network that provides
an informed initialization for the MPC procedure.

To this end, we train a policy model via behavior cloning from a small set of expert demonstrations.
Given a pretrained SAVi object-centric decomposition model, the policy network $\PolicyModel{\theta}$ is trained to predict the expert actions $\ActionT{t}$ from the corresponding structured slot-based latent representations $\SlotsT{t}$:
\vspace{-0.1cm}
\begin{equation}
	\Loss_{\PolicyModel{\theta}} =
		\sum_{t=1}^{\NumFrames}
		|| \PolicyModel{\theta}(\SlotsT{t}) - \ActionT{t}||_2^2.
\end{equation}
While such a policy may have limited generalization capabilities, we hypothesize that it can provide a strong initial action proposal, enabling more efficient MPC optimization and faster convergence.

\subsection{Model Predictive Control}
\label{sec:mpc}

At inference time, we perform model predictive control (MPC) in the learned latent
object-centric space in order to optimize actions sequences for reaching a goal state.
Given a goal image, we first encode it using the object-centric encoder
to obtain a set of latent goal slots $\SlotsT{\text{Goal}}$ representing the target object configuration.
Starting from the current observation at time step $t$, \method
encodes the image into latent object slots $\SlotsT{t}$, which serve
as the initial state for planning.
The learned dynamics model autoregressively forecasts future latent slot states conditioned on the initial slot representations and a candidate action sequence $\Actions{t}{t+H-1}$,
producing the predicted slot configuration $\PredSlotsT{t+H}$ at planning horizon $H$.

MPC optimizes the action sequence by minimizing the distance between
the last predicted latent slot configuration $\PredSlotsT{t+H}$ and the latent goal slots $\SlotsT{\text{Goal}}$.
Formally, the MPC objective is defined as
\begin{equation}
	\label{eq:mpc-objective}
	\mathcal{J}_{\text{MPC}}= ||\PredSlotsT{t+H}	- \SlotsT{\text{Goal}} ||_2^2 .
\end{equation}
At each optimization iteration, MPC evaluates candidate action sequences by rolling them out through the learned dynamics model, updates them based on their predicted costs, and executes only the first action of the optimized sequence before replanning at the next time step.

Since the ordering of object slots is not guaranteed to be consistent across time steps or between predicted and goal states for object-centric models, we apply Hungarian matching when computing MPC costs to ensure a meaningful comparison between object-centric latent states. At each time step, predicted object slots are aligned to goal slots by minimizing the pairwise Euclidean distance in latent space. The MPC costs are then computed after alignment, using the matched object representations.

We evaluate two different MPC variants, namely \emph{MPPI}~\citep{williams_mppi_2015} and \emph{gradient-based MPC}.
Following~\citet{hansen_td-mpc2_2024}, we additionally leverage a learned policy network to warm-start the optimization by providing good initial action trajectories. The policy is rolled out over the planning horizon using the dynamics model.

\paragraph{MPPI:}
MPPI is a sampling-based MPC method that iteratively updates parameters of a time-dependent multivariate Gaussian with diagonal covariance using importance-weighted trajectory costs.
We follow the variant used by~\citet{hansen_td-mpc2_2024}.
At optimization iteration $j$, the proposal distribution over action sequences is parameterized by
$(\mu^{j}, \sigma^{j})_{t:t+H-1}$, where $\mu^{j}_{t}, \sigma^{j}_{t} \in \mathbb{R}^{m}$ denote the distribution parameters for time step $t$ in the planning horizon $H$.
MPPI independently samples $N$ trajectories $\ActionT{t} \sim \mathcal{N}(\mu^{j-1}_{t}, (\sigma^{j-1}_{t})^{2} \mathrm{I})$ using rollouts generated by the learned model $d_{\theta}$, which are then estimated using the MPC objective defined in \cref{eq:mpc-objective}.

At iteration $j$, MPPI selects the top-$k$ trajectories with lower planning cost, and updates the proposal distribution parameters $\mu^{j}$ and $\sigma^{j}$ using an importance-weighted empirical estimate:
\begin{align}
    \label{eq:mppi-objective-norm}
    &\mu^{j} = \frac{\sum_{i=1}^{k} \Omega_{i} \Gamma_{i}^{\star}}{\sum_{i=1}^{k} \Omega_{i}}\,,~
    \sigma^{j} = \sqrt{\frac{\sum_{i=1}^{k} \Omega_{i} (\Gamma_{i}^{\star} - \mu^{j})^{2}}{\sum_{i=1}^{k} \Omega_{i}}}\,,
\end{align}
where $\Omega_{i} = e^{-\tau (\mathcal{J}_{\Gamma,i}^{\star})}$, $\tau$ is a temperature parameter controlling the sharpness of the weighting,
and $\Gamma_{i}^{\star}$ denotes the $i$th top-$k$ trajectory corresponding to cost estimate $\mathcal{J}_{\Gamma}^{\star}$.
After a fixed number of iterations $J$, the planning procedure is terminated and a trajectory is sampled from the final action proposal distribution.
We plan at each decision step $t$ and execute only the first action, i.e., we employ \emph{receding-horizon} MPC to produce a feedback policy.
To reduce the number of iterations required for convergence, we ``warm start'' trajectory optimization at each step $t$ by reusing the one-step shifted mean $\mu$ obtained at the previous step~\citep{argenson_model-based_2021}, but always use a large initial variance to avoid local minima.

\paragraph{Gradient-based MPC:}
A differentiable learned dynamics model enables computing gradients of the planning objective with respect to the action sequence by backpropagating through the rollout,
enabling gradient-based MPC to directly optimize a single candidate action sequence instead of sampling hundreds of trajectories as in MPPI.
Specifically, the actions are optimized by minimizing the planning objective (\cref{eq:mpc-objective}) via gradient descent:
\begin{equation}
	\Action \leftarrow \Action - \eta \mathbf{\nabla} \mathcal{J}_{\text{MPC}}.
\end{equation}

Inspired by~\citet{spieler2026dreammpc}, we use a policy network to efficiently guide the MPC procedure, which has shown to be beneficial for stabilizing gradient-based MPC.
In contrast to model-based RL methods, we do not jointly train the policy and world model, but instead learn each component independently from offline datasets without reward signals.
Rather than relying on learned reward or value functions, we directly
optimize the object-wise distance between predicted and goal object slots using latent-space distances as the planning objective.
This alternative formulation enables visual goal-directed planning without requiring rewards during training or planning.
Additionally, instead of sampling multiple trajectories from the policy network, \method directly optimizes a single trajectory initialized by the policy, thereby improving optimization efficiency.

\section{Experiments}
\label{sec:experiments}

To evaluate the effectiveness of our object-centric planning framework, we investigate the following research questions:
(i) Can an object-centric world model be learned purely from pre-collected offline trajectories and subsequently be used for goal-conditioned planning?
(ii) Do object-centric representations enable more efficient planning compared to holistic scene representations?
(iii) What components are required to efficiently solve long-horizon manipulation tasks?
We evaluate our proposed method on four robotic manipulation environments and compare against state-of-the-art world models to answer these questions.

\subsection{Evaluation Setup}

\paragraph{Datasets:}
We evaluate our proposed approach on four robotic manipulation environments adapted from two different benchmarks: \emph{Button Press} and \emph{Lever Pull} from \MetaWorld~\citep{yu2020meta}, and \emph{Stack} and \emph{Square} from \Robosuite~\citep{robosuite2020}.
For each environment, we generate two offline datasets: (i) trajectories collected by a random exploration agent, and (ii) a small set of expert demonstrations solving the corresponding task.
The random exploration dataset is used to train both the object-centric scene parsing model and the structured dynamics model, whereas the expert demonstrations are used to train the behavior cloning policy that warm-starts MPC.
We use only visual observations in all environments and do not rely on additional inputs
such as proprioceptive states.
Further details are provided in \cref{app:environments}.

\paragraph{Baselines:}
We compare \method against established baselines for both online and offline reinforcement learning, including goal-conditioned behavior cloning (GC-BC) \citep{lynch2020a,ghosh2021learning}, \mbox{Dreamer-v3}~\citep{hafner_mastering_2025}, and DINO-WM~\citep{zhou_dino-wm}.
DINO-WM learns a world model that operates in the latent space of a pretrained DINOv2~\citep{oquab2024dinov} encoder using offline, reward-free data.
The learned dynamics model is then used for planning via CEM, a gradient-free, sampling-based MPC technique.
Goal-conditioned behavior cloning (GC-BC) learns a goal-conditioned policy from reward-free, offline data via behavior cloning.
For a fair comparison, GC-BC, DINO-WM and \method are trained using the same offline datasets.
Dreamer-v3 is a widely used model-based RL method that jointly learns a visual world model and a policy from reward signals obtained through interaction with the environment.
Since this method requires rewards for training, we do not use the offline datasets, but instead follow its intended training protocol and learn the policy through online environment interactions.
Further details are provided in \cref{app:baselines}.

\paragraph{Success definition:}
For a fair comparison, all methods are evaluated using the same protocol.
At the beginning of each evaluation episode, the simulator state is initialized using the first state of an evaluation trajectory.
DINO-WM, GC-BC and \method additionally receive a goal image depicting successful task completion in the current environment.
At each time-step, all methods receive the current visual observation and select an action according to their policy.
The actions are executed in the environment and the subsequent observation is returned.
This procedure is repeated until either the goal is successfully reached or the maximum episode length is exceeded.
Our evaluation protocol differs from the procedure used by DINO-WM, which only considers short randomly sampled sub-trajectories. Instead, we evaluate full episodes, which better reflects long-horizon planning performance and task completion.
We report success rates across 50 evaluation episodes.
Results obtained using the original DINO-WM evaluation protocol are provided in \cref{sec:dino-wm-eval}.

\subsection{Comparison to Baselines}
A quantitative comparison of the different methods is presented in \cref{table:results}.
Our proposed method matches the performance of Dreamer-v3 on \MetaWorld{}, and outperforms all baselines on \Robosuite.
We further evaluate the quality of the predicted observations from DINO-WM and \method using PSNR, SSIM~\citep{Wang_SSIM_2004} and LPIPS~\citep{Zhang_TheUnreasonableEffectivenessOfDeepFeaturesLPIPS_2018}.
Although DINO-WM produces visually plausible predictions, as indicated in \cref{tab:image-metrics-comparison}, 
task completion completely fails for tasks that require planning over longer horizons.
Similar to \citet{wang2026temporalstraighteninglatentplanning} we find that the imagined rollouts for long horizons do not match the real dynamics.
Increasing the number of candidate trajectories and optimization iterations did not help to overcome this issue, likely due to the large search space.
We further include results for the original evaluation procedure used for DINO-WM in \cref{table:dino-wm-eval}, which show that the success rate in this setting also significantly decreases when removing proprioceptive information and considering longer horizons.
For more complex tasks such as \Robosuite{} Stack, DINO-WM fails to achieve goal-condition task success even for short planning horizons.

We hypothesize that this limitation arises because holistic latent representations would require substantially higher state-action coverage compared to object-centric representations, which introduce a compositional inductive bias. 
As a result, our model exhibits strong zero-shot generalization capabilities and overcomes limitations 
commonly observed in offline-trained policies, which generalize poorly and rely heavily on 
high-quality data and broad state-action coverage~\citep{Sobal2025Learning}.

\begin{table}[tpb]
	\caption{
		Task success comparison of \method with state-of-the-art baselines.
		\method matches Dreamer-v3 on Meta-World (\emph{Button Press} and \emph{Lever Pull}) and outperforms all baselines on \Robosuite{} (\emph{Stack} and \emph{Square}).
		Best two results are highlighted boldface and underlined, respectively. Wilson 95\% confidence intervals  are shown in brackets.
	}
	\label{table:results}
	\small
    \begin{center}
	\begin{threeparttable}
	\begin{tabular}{p{2cm} C{2.35cm}C{2.35cm}C{2.35cm}C{2.35cm}}
		\toprule
		\multicolumn{1}{c}{} &  \multicolumn{4}{c}{\textbf{Success Rates $\uparrow$}} \\
		\cmidrule(r){2-5} 
		\textbf{Method} & \textbf{Button Press} & \textbf{Lever Pull} & \textbf{Stack} & \textbf{Square} \\
		\midrule
		\method & \textbf{0.64} \footnotesize[0.50, 0.76] & \underline{0.52} \footnotesize[0.39, 0.65]  & \textbf{0.42} \footnotesize[0.29, 0.56] & \textbf{0.22} \footnotesize[0.13, 0.35] \\
		DINO-WM & 0.00 \footnotesize[0.00, 0.07] & 0.00 \footnotesize[0.00, 0.07] & 0.00 \footnotesize[0.00, 0.07] & 0.00 \footnotesize[0.00, 0.07] \\
		Dreamer-v3 & \textbf{0.64} \footnotesize[0.50, 0.76] & \textbf{0.56} \footnotesize[0.42, 0.69] & \underline{0.30} \footnotesize[0.19, 0.44] & 0.00 \footnotesize[0.00, 0.07] \\
		GC-BC & \underline{0.54} \footnotesize[0.40, 0.67] & 0.10 \footnotesize[0.04, 0.21] & \underline{0.30} \footnotesize[0.19, 0.44] & 0.00 \footnotesize[0.00, 0.07]  \\
		\bottomrule
	\end{tabular}
	\end{threeparttable}
    \end{center}
\end{table}
\begin{table}[t!]
	\centering
	\vspace{-0.4cm}
	\caption{Comparison of DINO-WM and \method across different environments on image metrics.}
	\label{tab:image-metrics-comparison}
	\resizebox{\textwidth}{!}{\begin{threeparttable}
	\setlength{\tabcolsep}{2pt}
	\begin{tabular}{l ccc ccc ccc ccc}
		\toprule
		& \multicolumn{3}{c}{\textbf{Button Press}} 
		& \multicolumn{3}{c}{\textbf{Lever Pull}} 
		& \multicolumn{3}{c}{\textbf{Stack}}
		& \multicolumn{3}{c}{\textbf{Square}} \\
		\cmidrule(lr){2-4} 
		\cmidrule(lr){5-7} 
		\cmidrule(lr){8-10} 
		\cmidrule(lr){11-13} 
		\textbf{Method} 
		& \textbf{PSNR}\scriptsize{$\uparrow$}
		& \textbf{SSIM}\scriptsize{$\uparrow$}
		& \textbf{LPIPS}\scriptsize{$\downarrow$}
		& \textbf{PSNR}\scriptsize{$\uparrow$}
		& \textbf{SSIM}\scriptsize{$\uparrow$}
		& \textbf{LPIPS}\scriptsize{$\downarrow$}
		& \textbf{PSNR}\scriptsize{$\uparrow$}
		& \textbf{SSIM}\scriptsize{$\uparrow$}
		& \textbf{LPIPS}\scriptsize{$\downarrow$}
		& \textbf{PSNR}\scriptsize{$\uparrow$}
		& \textbf{SSIM}\scriptsize{$\uparrow$}
		& \textbf{LPIPS}\scriptsize{$\downarrow$}\\
		\midrule
		DINO-WM & 35.84 & \textbf{0.990} & 0.0084 & \textbf{25.12} & \textbf{0.873} & \textbf{0.0393} & 26.90 & 0.964 & 0.0313 & 23.38 & \textbf{0.938} & 0.0418 \\
		\method & \textbf{35.92} & 0.977 & \textbf{0.0037} & 23.83 & 0.842 & 0.0426 & \textbf{32.85} & \textbf{0.980} & \textbf{0.0146} & \textbf{25.51} & 0.919 & \textbf{0.0354} \\
		\bottomrule
	\end{tabular}
	\end{threeparttable}}
\end{table}

\subsection{Model Analysis}

We conduct ablation studies to isolate the contribution of the different design choices in \method, including the object-centric world model, the policy prior, and the MPC formulation.
The results in \cref{table:ablation model} show that the object-centric world model, the learned policy network and gradient-based MPC are the most critical components, as removing them leads to the largest performance drops.
In particular, we find that tasks requiring long planning horizons (e.g. those from \Robosuite) strongly benefit from a good policy prior to guide the optimization process through the search space.

Our experiments further show that gradient-based MPC is substantially more effective than sampling-based MPC 
methods such as MPPI in the considered offline setting, where state-action coverage is limited.
Even when using the policy prior to guide the sampling procedure, the performance deteriorates for MPPI significantly.
We hypothesize that this is due to the dynamics model being queried out of distribution, leading to suboptimal optimization.
In this setting, the characteristics of gradient-based MPC, which stays closer to the nominal policy distribution, are more beneficial than the higher diversity of sampling-based MPC methods.

For \MetaWorld, \method often moves visually close to the goal, but does not always apply the last bit of force or displacement needed to fully press the button, or misses or slips off the lever.
This is particularly distinct without a policy prior, and is also the reason for minor or no performance improvement of MPC compared to using the policy only.
Adding proprioceptive information or using subgoals could help to mitigate this and further improve performance. 
For the more complex tasks from \Robosuite, gradient-based MPC improves performance notably compared to the policy only, highlighting the importance of planning for long-horizon manipulation tasks.

\input{tables/ablations.tex}

We also perform an ablation study of the MPC objective by using different formulations of slot-based cost functions.
The results, summarized in \cref{table:ablation-mpc-objective}, show that performing MPC with a cost function using the sum of squared distances (SSE) or cosine similarity of slots performs best.
Using the slot masks $\ObjectMaskNorm$ is biased towards having accurate modeling of the background and larger objects instead of small objects.
While this bias can be mitigated by using the normalized intersection over union (IoU) between slot masks, directly optimizing over slot representations is preferable, since the same representation space is also used during predictor training.

Since inference time is a critical factor when deploying a model to a real-world system such as a robot, in \cref{table:inference} we report the time required for a single planning step on a single NVIDIA RTX A6000 GPU.
Object-centric slot representations substantially reduce the latent feature space compared to patch-based methods, such as DINO-WM, reducing the representation 
size from \texttt{\#Tokens$\times$$\DimFeats$} to \texttt{$\NumSlots\times\SlotDim$}.
In our setting, we use four slots with a dimensionality of 128, resulting in a latent space of 
$4\times128$ compared to $196\times384$ for DINO-WM.
This corresponds to an approximately 99\% reduction in latent dimensionality and leads to significantly faster 
planning times, even when using sampling-based MPPI.
Additionally, warm-starting MPC by using a policy prior further improves computational efficiency of \method.

\begin{table}[htpb]
	\caption{
		Comparison of planning execution times for different methods on a single NVIDIA RTX A6000 GPU.
		\method is substantially more computationally efficient than the patch-based baseline.
	}
	\label{table:inference}
	\small
    \begin{center}
	\begin{threeparttable}
	\begin{tabular}{p{4.75cm} C{2.75cm}C{2.75cm}}
		\toprule
		\multicolumn{1}{c}{} &  \multicolumn{2}{c}{\textbf{Planning time (s) $\downarrow$}} \\
		\cmidrule(r){2-3} 
		\textbf{Method} & \textbf{\MetaWorld} & \textbf{\Robosuite} \\
		\midrule
		\method & \textbf{0.42 $\pm$ 0.01} & \textbf{0.48 $\pm$ 0.02 }\\
		\hspace{0.2cm} w/ MPPI (no gradient-based MPC) & \underline{4.22 $\pm$ 0.06} & \underline{5.19 $\pm$ 0.03} \\
		DINO-WM & 144.37 $\pm$ 0.83 & 145.30 $\pm$ 0.72 \\
		\bottomrule
	\end{tabular}
	\end{threeparttable}
    \end{center}
    \vspace{-0.3cm}
\end{table}

\section{Conclusion}
\label{sec:conclusion}

We study the importance of object-centric representations for efficient planning with MPC and introduce \method, a novel method for performing gradient-based MPC with object-centric latent dynamics.
\method learns object-centric representations from reward-free offline data and combines a slot-based 
world model with gradient-based trajectory optimization for goal-conditioned planning directly from visual 
observations.
By optimizing action trajectories in a compact object-centric  latent space, \method significantly improves both task 
performance and computational efficiency compared to state-of-the-art baselines.
Our experiments on simulated robotic manipulation tasks demonstrate that object-centric representations are a 
powerful inductive bias for control, particularly for long-horizon planning problems.
Overall, our results suggest that structured object-centric world models are a promising direction 
for scalable and efficient model-based control from raw visual observations.

\section*{Acknowledgements}
This work was funded by the Federal Ministry of Research, Technology and Space of Germany (BMFTR) within the WestAI - AI Service Center West, grant no. 16IS22094A and within the Robotics Institute Germany, grant no. 16ME0999. Computational resources were provided by the German AI Service Center WestAI.

\bibliographystyle{plainnat}
\bibliography{references}

\newpage

\appendix
    \hrule height 3pt
    \vskip 3mm
    \begin{center}
        \Large{\textbf{Appendix}}
    \end{center}
    \vskip 3mm
    \hrule height 1pt
\vspace{5mm}
\startcontents[sections]
\printcontents[sections]{l}{1}{\setcounter{tocdepth}{2}}
\vskip 8mm
\hrule height .5pt
\vskip 10mm
\startcontents

\section{Limitations and Future Work}\label{app:limitations}

\method relies like DINO-WM on goal images for planning and on ground truth actions for training the dynamics model and policy, which both may not always be available. An extension of this work could involve text conditioning \citep{villar-corrales2026textocvp}, which allows for specifying the goal using natural language. Performance depends on the quality of the object-centric decomposition model and the policy prior. Further improving performance by using subgoals and extending the framework to real-world robotic settings is part of future work, potentially leveraging more capable decomposition models such as DINOSAUR~\citep{Seitzer_BridgingTheGapToRealWorldObjectCentricLearning_2023}.

\section{Datasets and Simulation Environments}\label{app:environments}

\begin{figure}[pb]
    \captionsetup[subfigure]{labelformat=empty}
    \begin{subfigure}[ht]{0.25\textwidth}
        \centering
        \frame{\includegraphics[width=0.75\textwidth]{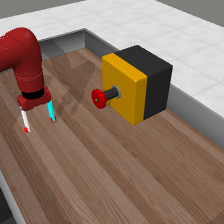}}
        \vspace{-0.2em}
        \caption{Button Press}
    \end{subfigure}%
    \hfill
	\begin{subfigure}[ht]{0.25\textwidth}
        \centering
        \frame{\includegraphics[width=0.75\textwidth]{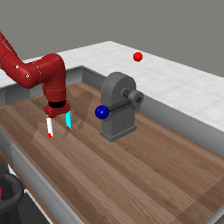}}
        \vspace{-0.2em}
        \caption{Lever Pull}
    \end{subfigure}%
    \hfill
	\begin{subfigure}[ht]{0.25\textwidth}
        \centering
        \frame{\includegraphics[width=0.75\textwidth]{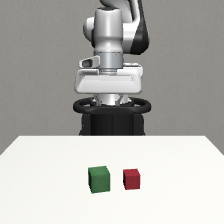}}
        \vspace{-0.2em}
        \caption{Stack}
    \end{subfigure}%
    \hfill
    \begin{subfigure}[ht]{0.25\textwidth}
        \centering
        \frame{\includegraphics[width=0.75\textwidth]{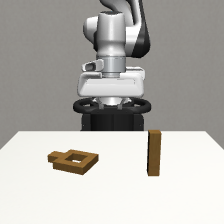}}
        \vspace{-0.2em}
        \caption{Square}
    \end{subfigure}%
    \caption{
        \textbf{Environments.} We evaluate \method on four different environments.
    }
    \label{fig:environments}
\end{figure}

We evaluate \method and baseline models on four tasks from different manipulation benchmarks.
Image inputs are resized to a resolution of 64$\times$64 for \method and Dreamer-v3, and 224$\times$224 for DINO-WM.
The environments are illustrated in \cref{fig:environments}.

\paragraph{\MetaWorld~\citep{yu2020meta}:}
is an open source benchmark (MIT license) containing continuous control robotic manipulation environments.
We consider the \emph{Button Press} task, which requires the robot to press a button that is randomly positioned in the scene, and the \emph{Lever Pull} task, where the robot needs to pull a lever whose position is randomly initialized.
All tasks from \MetaWorld{} share the same embodiment, observation space (\texttt{dim(S) = 39}) and action space (\texttt{dim(A) = 4}).
We refer to~\citet{yu2020meta} for the definitions of the reward functions and success metrics used in the \MetaWorld{} tasks.
We generate a training dataset consisting of \num{9000} training sequences and \num{1000} validation trajectories using a random exploration policy.
We use the provided expert policies from \MetaWorld{} to generate a small training set of 200 expert demonstrations, as well as an evaluation dataset containing 50 successful sequences.

\paragraph{\Robosuite~\citep{robosuite2020}:}
We further consider the \emph{Stack} task from \emph{\Robosuite}, 
where the goal is to stack two blocks on top of each other. The blocks are randomly positioned on the table.
Additionally, we consider the \emph{Square\_D1} task from \emph{\MimicGen}, where the robot needs to pick a square nut and place it on a rod. The positions of the nut and rod are randomly initialized.
The action dimension of the used Panda robot is seven.
We generate a dataset containing \num{10000} sequences, split into \num{9000} training and \num{1000} validation sequences using a random exploration policy.
We use \emph{\MimicGen}~\citep{mimicgen_2023} to generate \num{2000} expert demonstrations for behavior cloning and evaluation, which are split into \num{1800} sequences for training the policy and 200 sequences for evaluation.
\MimicGen{} requires using an \texttt{OSC\_POSE} controller, which controls the end-effector pose.
\Robosuite{} is licensed under an MIT license and \MimicGen{} under an NVIDIA license.

\section{Implementation Details}

In this section, we describe the network architecture and training details for each of the components of \method.
Our models are implemented in PyTorch~\citep{Paszke_AutomaticDifferneciationInPytorch_2017} and are trained on a single NVIDIA RTX A6000 GPU.

\subsection{Object-Centric Learning and World Modeling}
We closely follow~\citet{villar_PlaySlot_2025} for the implementation of both the 
object-centric decomposition model and the structured dynamics model.

\paragraph{Object-Centric Decomposition:}
The object-centric decomposition is based on SAVi~\citep{Kipf_ConditionalObjectCentricLearningFromVideo_2022}, a recursive slot-based model that serves as our scene parsing and object rendering modules.
Specifically, we adopt their proposed CNN-based image encoder $\ImageEncoder$ and slot decoder $\SlotDecoder$, as well as their transformer-based transition module, and Slot Attention corrector.
We use four 128-dimensional object slots for all the evaluated dataset, which suffice to separate the agent, the different objects, and the background. 
For all datasets, the initial slot representations $\SlotsT{0}$ are randomly initialized and optimized as learnable parameters via backpropagation.
Furthermore, we use three Slot Attention iterations on the first observation to ensure a stable initial decomposition. For subsequent frames, a single iteration suffices to recursively refine the slot representations conditioned on the newly observed image features.

\paragraph{Object-Centric World Modeling:}

Our cOCVP structured world model is an object-centric
transformer predictor inspired by
\citet{Villar_OCVP_2023,Wu_SlotFormer_2023}.
The cOCVP module consists of four transformer layers with 256-dimensional tokens, eight attention heads of dimension 64, and a feed-forward hidden dimension of 1024.

To enable action-conditioned prediction, cOCVP maps
both the actions $\Actions{1}{t}$ and object slots $\SlotsT{1:t}$ into a shared token embedding space using learnable projection layers.
The projected object slots are then conditioned by adding the corresponding projected action at each time-step.
Furthermore, following~\citet{Wu_SlotFormer_2023}, we augment the tokens with a temporal sinusoidal positional embedding, which assigns the same encoding to all tokens from the same time-step, thus preserving the inherent permutation equivariance of the objects.

\subsection{Policy Model}
The policy model $\PolicyModel{\theta}$ is a transformer that jointly processes the objects slots from a single time step $\SlotsT{t}$ together with an additional learnable action embedding \ActToken in order to regress an expert action.
The model consists of four transformer layers with 256-dimensional tokens, four 64-dimensional heads and a feed-forward hidden dimension of 1024.
Through the attention mechanism, information from the object slots is aggregated into the \ActToken token, which is subsequently mapped to produce a single action $\PredAction$ using a learnable linear projection head.
The policy is rolled out autoregressively over the horizon $H$ using the learned dynamics model to generate an initial action sequence for MPC.

\subsection{Training Details}

\paragraph{SAVi Training:}  SAVi is trained for object-centric decomposition using the Adam optimizer~\citep{kingma_adam_2015}, a batch size of 64, sequences of length eight frames, and a base learning rate of $10^{-4}$, which is linearly warmed-up for the first 4000 steps, followed by cosine annealing for the remaining of the training process.
Moreover, we clip the gradients to a maximum norm of 0.05.

\paragraph{cOCVP~Training:} We train our cOCVP module given a pretrained SAVi decomposition model.
This module is trained with the Adam optimizer~\citep{kingma_adam_2015}, batch size of 64, and a base learning rate of $2\times10^{-4}$, which decreases during training with a cosine annealing schedule.
To stabilize the training, we clip the gradients to a maximum norm of 0.05.
We set the loss weights to $\lambda_{\text{Img}} = 1$, and $\lambda_{\text{Slot}} = 1$.

\paragraph{$\PolicyModel{\theta}$~Training:} We train the $\PolicyModel{\theta}$ module given pretrained and frozen SAVi, and cOCVP modules.
This module is trained with the Adam optimizer~\citep{kingma_adam_2015}, batch size of 64, and a learning rate of $3\times10^{-4}$.

\subsection{Model Predictive Control}
We compare two different MPC methods: gradient-based MPC and MPPI. For both, we use a policy network to warm-start the optimization and clip actions to the valid actions bounds. 

\begin{table}[htpb]
	\caption{
		Gradient-based MPC parameters.
	}
	\label{table:gradient-based-mpc-params}
    \begin{center}
	\begin{tabular}{ll}
		\toprule
		\textbf{Hyperparameter} &  \textbf{Value} \\
		\midrule
		Horizon $\Horizon$ & 8 (Square, \MetaWorld) \\
		& 15 (Stack) \\
		Iterations & 3 \\
		Number of samples & 1 \\
		Policy prior samples & 1 \\
		Step size $\eta$ & 0.001 \\
		\bottomrule
	\end{tabular}
    \end{center}
\end{table}

\begin{table}[htpb]
	\caption{
		MPPI parameters.
	}
	\label{table:mppi-params}
    \begin{center}
	\begin{tabular}{ll}
		\toprule
		\textbf{Hyperparameter} &  \textbf{Value} \\
		\midrule
		Horizon $\Horizon$ & 15 \\
		Iterations & 5 \\
		Number of samples & 64 \\
		Number of elites & 16 \\
		Policy prior samples & 16 \\
		Minimum std. & 0.05 \\
		Maximum std. & 2.0 \\
		Temperature & 1.0 \\
		\bottomrule
	\end{tabular}
    \end{center}
\end{table}

\section{Baselines}
\label{app:baselines}
\paragraph{Dreamer-v3:} For Dreamer-v3, we use the official reimplementation from \url{https://github.com/danijar/dreamerv3}, which is licensed under an MIT license. For a fair comparison, we do not use the offline dataset, but train the agent through interaction with the environment for 1M steps for \MetaWorld, and 5M steps for \Robosuite. We follow the authors' suggested hyperparameters for visual observations (DM Control) and use an update-to-data (UTD) ratio of 256 for all environments. We use a model size of 12M parameters for \MetaWorld{} and 25M parameters for \Robosuite{}. Please refer to \citet{hafner_mastering_2025} for a complete list of hyperparameters.

\paragraph{DINO-WM:} For DINO-WM, we use the official implementation provided by the authors and licensed under an MIT license: \url{https://github.com/gaoyuezhou/dino_wm}. We use the default hyperparameters suggested by the authors.

\paragraph{OCVP:} For our experiments, we use OCVP as the object-centric world model and base our implementation on the official implementation of PlaySlot \citep{villar_PlaySlot_2025}: \url{https://github.com/angelvillar96/PlaySlot}.

\paragraph{Non-Object-Centric Baseline:} This baseline model follows the same general framework as our proposed model, but replaces the object-centric SAVi encoder and decoder with a simple convolutional auto-encoder while keeping the other modules unchanged; thus allowing us to ablate the effect of object-centric representations for MPC compared to a single holistic latent.

\section{Additional Results}

\subsection{DINO-WM Evaluation Procedure}\label{sec:dino-wm-eval}
We further provide results when evaluating DINO-WM using the procedure used by the authors, i.e., sampling random subtrajectories of length $H=25$ from the evaluation dataset and using the last observation for planning. Success is defined as reaching a state that matches the goal state up to some threshold. We find that removing proprioceptive information leads to a significant decrease of the success rate. Our experiments show that with increasing subtrajectory length, the success rate significantly drops even when significantly increasing the number of candidate trajectories, aligning with a success rate of zero obtained when evaluating the complete trajectory following our evaluation procedure. Recent works also find that removing proprioceptive information \citep{terver2026drivessuccessphysicalplanning} and using a longer goal horizon \citep{parthasarathy2025closingtraintestgapworld, psenka2026parallelstochasticgradientbasedplanning,wang2026temporalstraighteninglatentplanning} both lead to a significant drop of the success rate for DINO-WM.

For \MetaWorld{} and \Robosuite, we consider a task as successful if the distance between the actual system state and the desired goal state (in the original state space) is less than a predefined threshold. This threshold is set to 0.3 for \MetaWorld{} and 1.0 for \Robosuite.

\begin{table}[htpb]
	\caption{
		DINO-WM evaluation results using the procedure proposed by the authors.
	}
	\label{table:dino-wm-eval}
    \begin{center}
	\begin{tabular}{p{4.25cm} C{1.75cm}C{1.5cm}C{1.25cm}C{1.25cm}}
		\toprule
		\multicolumn{1}{c}{} &  \multicolumn{4}{c}{\textbf{(Sub-) Goal Reaching Rate $\uparrow$}} \\ 
		\cmidrule(r){2-5} 
		\textbf{Model} & Button Press & Lever Pull & Stack & Square \\
		\midrule
		\method & \textbf{0.80} & \textbf{0.86} & \textbf{0.38} & \textbf{0.18} \\
		DINO-WM & 0.56  & 0.10 & 0.00 & 0.04 \\
		\hspace{0.2cm} w/ goal horizon $H=50$ & 0.28 & 0.00 & 0.00 & 0.00 \\ 
		\bottomrule
	\end{tabular}
    \end{center}
\end{table}

\subsection{Qualitative Results}\label{sec:qualitative-results}
\begin{figure}[!ht]
  \setlength{\belowcaptionskip}{5pt}
  \centering
  \begin{subfigure}[b]{\linewidth}
	\centering
    \includegraphics[width=0.8\linewidth]{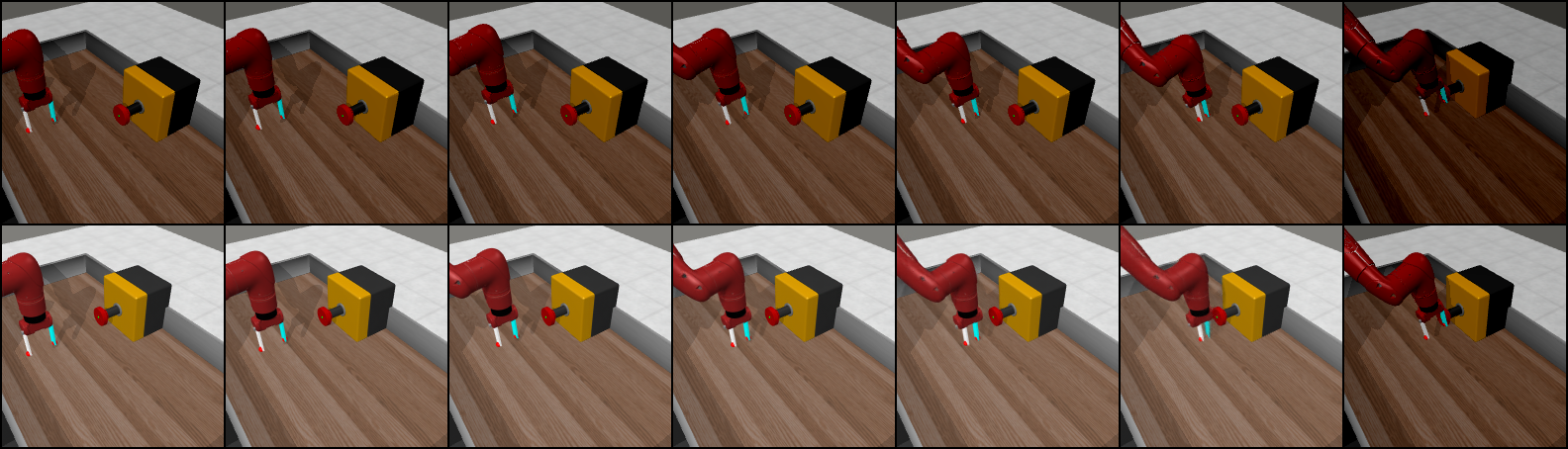}
    \caption{Goal horizon H=25}
	\label{fig:dino-wm-subtrajectories-gH25}
  \end{subfigure}
  \begin{subfigure}[b]{\linewidth}
	\centering
    \includegraphics[width=\linewidth]{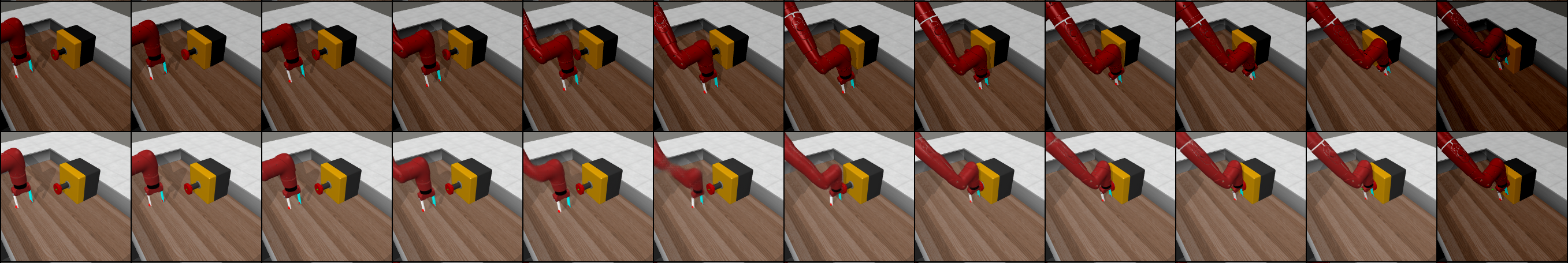}
    \caption{Goal horizon H=50}
	\label{fig:dino-wm-subtrajectories-gH50}
  \end{subfigure}
  \caption{
      \textbf{DINO-WM evaluation on subtrajectories.} (a) With a goal horizon of H=25. (b)  With a goal horizon of H=50. The bottom row corresponds to the predicted frames and the top row (shaded for visual distinction) are the actual observations from the simulator. The last image is the goal image.
  }
  \label{fig:dino-wm-subtrajectories}
\end{figure}

\cref{fig:dino-wm-subtrajectories} shows that while the predicted rollouts of DINO-WM are quite accurate for smaller goal horizons, the predictions deviate from the actual environment rollout for longer goal horizons. As shown in \cref{fig:dino-wm-subtrajectories-gH50}, DINO-WM predicts that the button is pressed (bottom row), while the robot actually misses the button by being too far right of it (top row).

\begin{figure}[!htpb]
  \centering
    \includegraphics[width=0.9\textwidth]{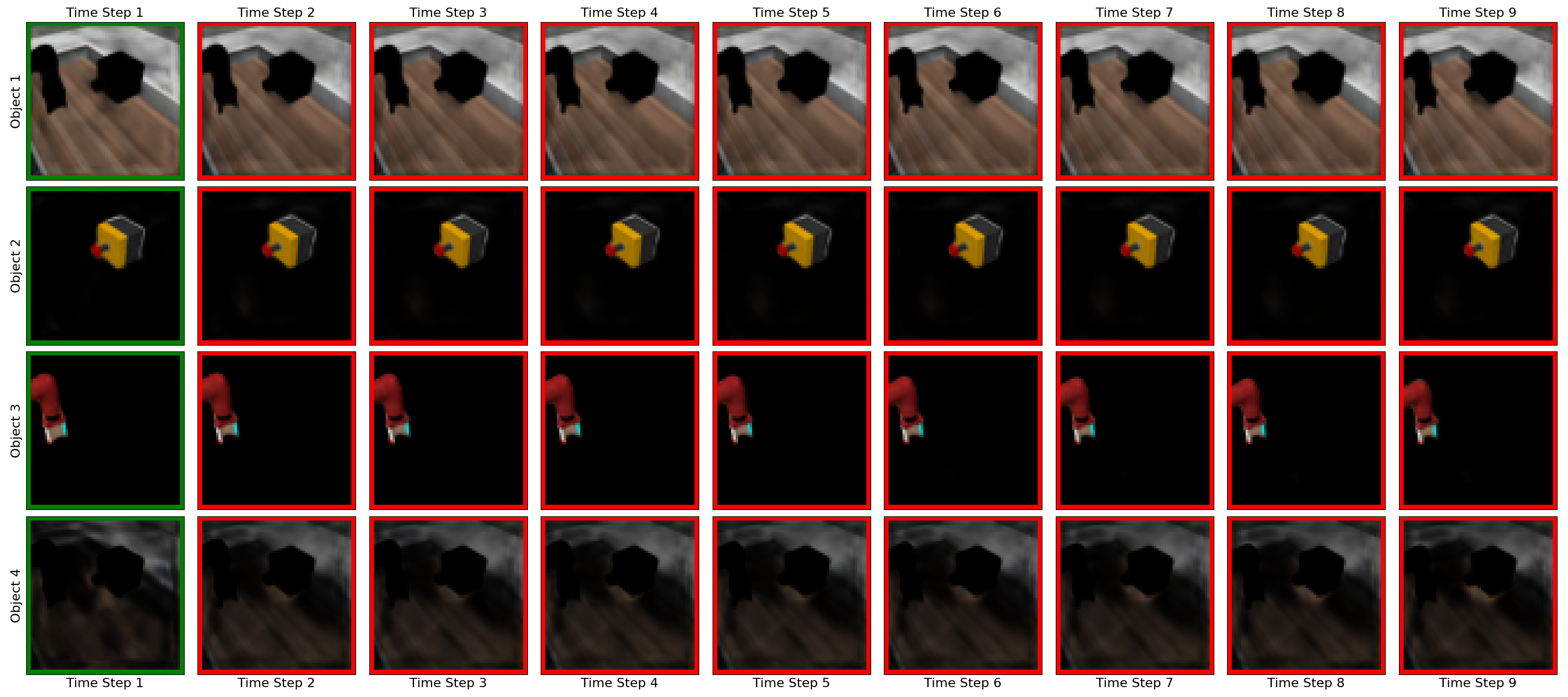}
  \caption{
      Predictions and decomposition results for Button Press. \method assigns a slot for the background, a slot for the robot arm, and a slot for the button.
  }
  \label{fig:mw-button-press-slots}
\end{figure}

\begin{figure}[!htpb]
  \centering
    \includegraphics[width=0.9\textwidth]{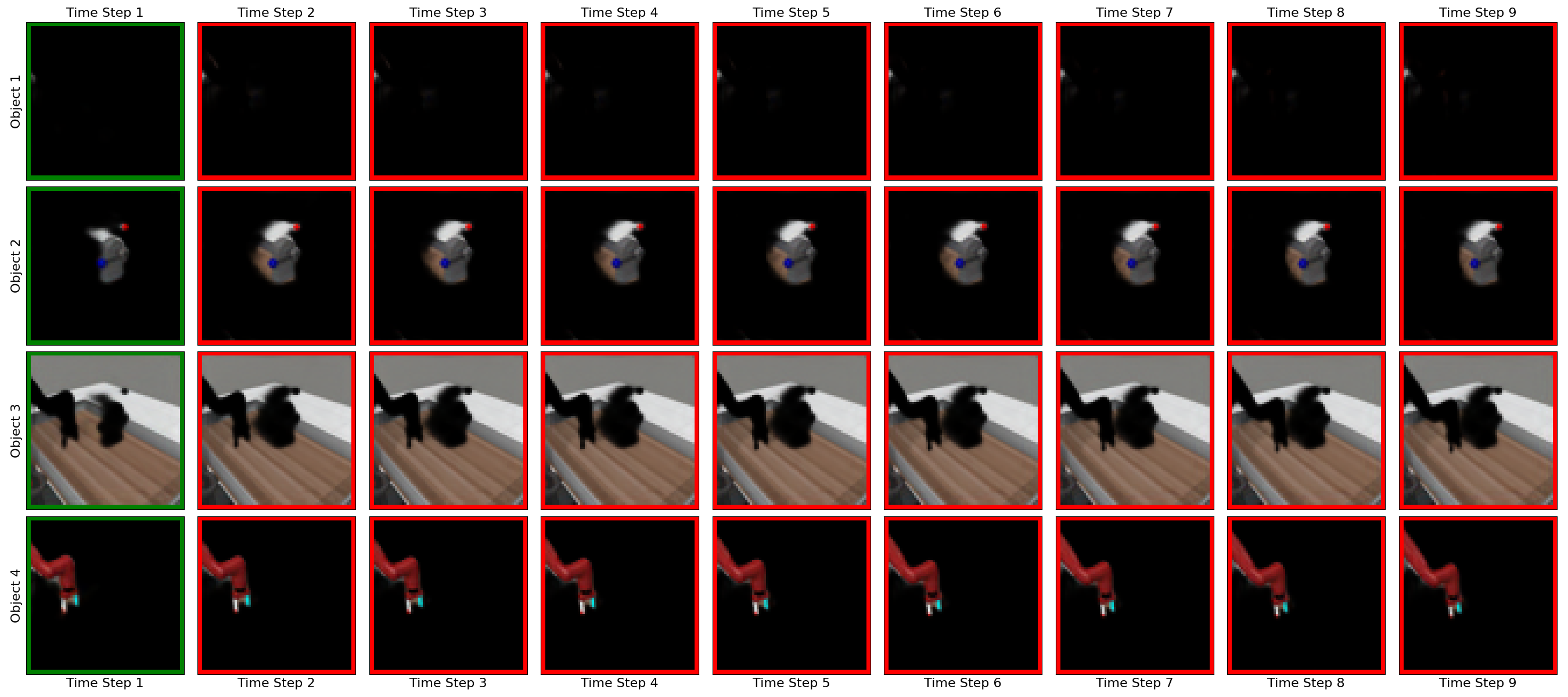}
  \caption{
      Predictions and decomposition results for Lever Pull. \method assigns a slot for the background, a slot for the robot arm, and a slot for the lever.
  }
  \label{fig:mw-lever-pull-slots}
\end{figure}

\begin{figure}[!htpb]
  \centering
    \includegraphics[width=0.9\textwidth]{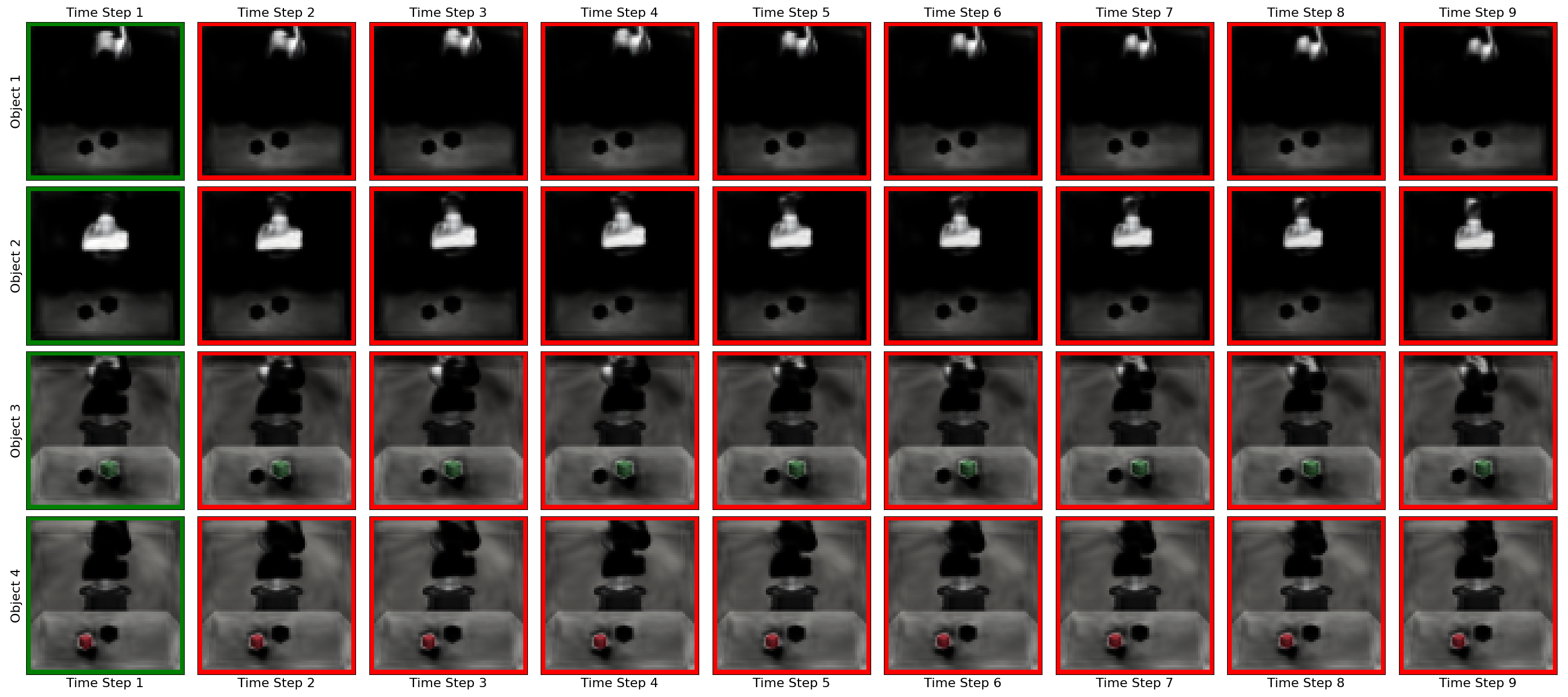}
  \caption{
      Predictions and decomposition results for Stack. \method assigns a slot for the background, a slot for the robot's gripper, and a slot for each cube.
  }
  \label{fig:robosuite-stack-slots}
\end{figure}

\begin{figure}[!htpb]
  \centering
    \includegraphics[width=0.9\textwidth]{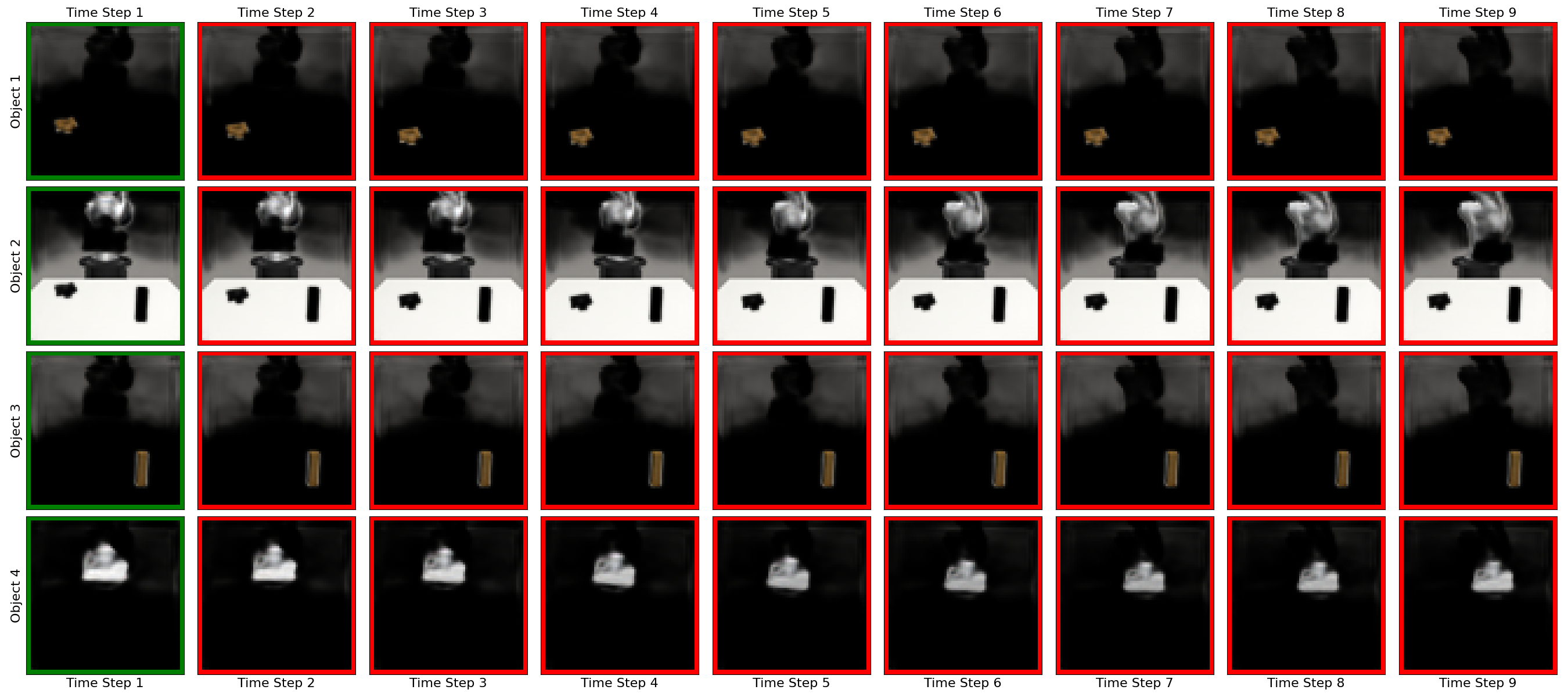}
  \caption{
      Predictions and decomposition results for Square. \method assigns a slot for the background, a slot for the robot's gripper, a slot for the peg, and a slot for the nut.
  }
  \label{fig:robosuite-square-slots}
\end{figure}

We visualize the predictions and decomposition results of the object-centric models of \method for the considered environments. \cref{fig:mw-button-press-slots,fig:mw-lever-pull-slots,fig:robosuite-stack-slots,fig:robosuite-square-slots} show that scenes are parsed into meaningful object representations.

\vspace{0.4in}
\begin{center}
    \textcolor{gray}{\textbf{\textemdash~Appendices continue on next page~\textemdash}}
\end{center}

\clearpage

\end{document}

%% file: macros.tex
\usepackage[utf8]{inputenc} 
\usepackage[T1]{fontenc}    

\usepackage{amsfonts}

\usepackage{amsmath}
\usepackage{amssymb}
\usepackage{mathtools}
\usepackage{mathrsfs}           
\usepackage{amsthm}
\theoremstyle{plain}

\theoremstyle{definition}

\theoremstyle{remark}

\usepackage{enumitem}
\usepackage{siunitx}
\usepackage{nicefrac}
\usepackage{microtype}
\usepackage{graphicx}
\usepackage{xcolor}         
\usepackage{booktabs}
\usepackage{array}
\usepackage{ragged2e}
\usepackage{etoolbox}
\usepackage{xspace}
\usepackage{float}
\usepackage{tabularx}
\usepackage{multirow}
\usepackage{threeparttable}
\usepackage{etoc} 
\usepackage{titletoc}
\usepackage{tcolorbox}
\usepackage{fancyvrb}
\usepackage{listings}
\usepackage{subcaption}
\usepackage[space]{grffile}     
\usepackage{balance}
\usepackage{hyperref}
\usepackage[capitalise,noabbrev,nameinlink]{cleveref}
\usepackage{url}

\newcommand{\method}{Slot-MPC\xspace}

\newcommand{\projectpage}{\url{https://slot-mpc.github.io}}

\Crefname{algorithm}{Alg.}{Algs.}
\Crefname{equation}{Eq.}{Eqs.}
\Crefname{figure}{Fig.}{Figs.}
\Crefname{tabular}{Tab.}{Tabs.}
\Crefname{table}{Tab.}{Tabs.}

\newcommand{\PreserveBackslash}[1]{\let\temp=\\#1\let\\=\temp}
\newcolumntype{C}[1]{>{\PreserveBackslash\centering}p{#1}}
\newcolumntype{R}[1]{>{\PreserveBackslash\raggedleft}p{#1}}
\newcolumntype{L}[1]{>{\PreserveBackslash\raggedright}p{#1}}
\newcolumntype{N}[1]{>{\centering\arraybackslash}m{#1}}

\definecolor{mygreen}{HTML}{7CB518}
\definecolor{myblue}{HTML}{3A86FF}
\definecolor{mypurple}{HTML}{8338EC}
\definecolor{mypink}{HTML}{FF006E}
\definecolor{myorange}{HTML}{FB5607}
\definecolor{myyellow}{HTML}{FFBE0B}

\definecolor{codegreen}{rgb}{0,0.5,0}
\definecolor{codered}{rgb}{0.7,0.1,0.1}
\definecolor{codegray}{rgb}{0.5,0.5,0.5}
\definecolor{codepurple}{rgb}{0.58,0,0.82}
\definecolor{backcolour}{rgb}{1,1,1}
\lstdefinestyle{python}{
    language=Python,
    backgroundcolor=\color{backcolour},
    commentstyle=\color{codered}\textit,
    keywordstyle=\bfseries\color{codegreen},
    numberstyle=\tiny\color{codegray},
    stringstyle=\color{codepurple},
    basicstyle=\ttfamily\scriptsize,
    breakatwhitespace=false,
    breaklines=true,
    captionpos=b,
    keepspaces=true,
    numbers=left,
    numbersep=4pt,
    showspaces=false,
    showstringspaces=false,
    showtabs=false,
    tabsize=1,
    fancyvrb=true
}
\newtcolorbox{conclusionbox}{
  colback=blue!5,        %
  colframe=blue!75!black, %
  coltitle=black,        %
  fonttitle=\bfseries,   %
  boxrule=1pt,           %
  arc=1mm,               %
  left=2mm,              %
  right=2mm,             %
  top=1mm,               %
  bottom=1mm,            %
}

\newcommand{\Action}{\textbf{a}}
\newcommand{\ActionT}[1]{\textbf{a}_{#1}}
\newcommand{\Actions}[2]{\textbf{a}_{#1:#2}}
\newcommand{\PredAction}{\mathbf{\hat{a}}}

\newcommand{\ActionProj}{f_{\Action}}
\newcommand{\ActToken}{\texttt{[ACT]}}

\newcommand{\Horizon}{H}
\newcommand{\Objective}{\mathcal{J}}

\newcommand{\Loss}{\mathcal{L}}

\newcommand{\PolicyModel}[1]{\pi_{#1}}

\newcommand{\ImageEncoder}{\mathcal{E}_{\textrm{SAVi}}}
\newcommand{\SlotDecoder}{\mathcal{D}_{\text{SAVi}}}

\newcommand{\Slots}{\mathbf{S}}
\newcommand{\SlotDim}{D_{\Slots}}
\newcommand{\NumSlots}{N_\Slots}

\newcommand{\SlotsT}[1]{\textbf{S}_{#1}}
\newcommand{\SlotSingleT}[2]{\textbf{s}_{#1}^{#2}}
\newcommand{\PredSlotsT}[1]{\hat{\textbf{S}}_{#1}}

\newcommand{\FeatureMaps}{\textbf{h}}
\newcommand{\DimFeats}{D_h}
\newcommand{\NumLocs}{L}
\newcommand{\Attention}{\textbf{A}}
\newcommand{\ObjectImage}{\textbf{o}_t^n}
\newcommand{\ObjectMask}{\textbf{m}_t^n}
\newcommand{\ObjectMaskNorm}{\tilde{\mathbf{m}}_t^n}

\newcommand{\ImageT}[1]{\textbf{X}_{#1}}
\newcommand{\PredImageT}[1]{\hat{\textbf{X}}_{#1}}
\newcommand{\NumFrames}{T}
\newcommand{\NumPreds}{T}

\newcommand{\softmax}{\mathrm{softmax}}

\def\R{{\mathbb R}}

\newcommand{\MetaWorld}{Meta-World}
\newcommand{\PushT}{Push-T}
\newcommand{\Robosuite}{robosuite}
\newcommand{\MimicGen}{MimicGen}

%% file: tables/ablations.tex
\begin{table}[htpb]
	\centering
	\caption{
		Ablation studies \method.
		\textbf{(a)} Impact of removing key components, including object-centric representations, MPC, and policy 
		initialization, as well as replacing gradient-based MPC with MPPI.
		\textbf{(b)} Comparison of different MPC objectives.
		Best two results are bolded and underlined, respectively.
	}
	\label{table:ablations}
	\begin{subtable}{\textwidth}
		\vspace{-0.1cm}
		\caption{
			\method component ablation.
		}
		\vspace{-0.4cm}
		\centering
		\small
		\label{table:ablation model}
		\begin{center}
			\begin{threeparttable}
				\begin{tabular}{p{4.75cm} C{2cm}C{1.75cm}C{1.25cm}C{1.25cm}}
					\toprule
					\multicolumn{1}{c}{} &  \multicolumn{4}{c}{\textbf{Success Rates $\uparrow$}} \\
					\cmidrule(r){2-5} 
					\textbf{Model Variant} & \textbf{Button Press} & \textbf{Lever Pull} & \textbf{Stack} & \textbf{Square} \\
					\midrule
					\method & \textbf{0.64}  & \textbf{0.52} &  \textbf{0.42} & \textbf{0.22} \\
					\hspace{0.2cm} w/o object-centric representations & \underline{0.62} & \underline{0.48} & 0.20 & 0.04 \\
					\hspace{0.2cm} w/o MPC & \textbf{0.64} & \textbf{0.52} & \underline{0.36} & \underline{0.18}  \\ 
					\hspace{0.2cm} w/o policy (zeros as initial actions) & 0.32 & 0.18 & 0.00 & 0.00 \\ 
					\hspace{0.2cm} w/ MPPI (no gradient-based MPC) & 0.04  & 0.04 & 0.00 & 0.00 \\
					\bottomrule
				\end{tabular}
			\end{threeparttable}
		\end{center}
	\end{subtable}
	\begin{subtable}{\textwidth}
		\vspace{0.1cm}
		\caption{MPC objective ablation.}
		\vspace{-0.4cm}
		\centering
		\small
		\label{table:ablation-mpc-objective}
		\begin{center}
			\begin{threeparttable}
				\begin{tabular}{p{4.75cm} C{2cm}C{1.75cm}C{1.25cm}C{1.25cm}}
					\toprule
					\multicolumn{1}{c}{} &  \multicolumn{4}{c}{\textbf{Success Rates $\uparrow$}} \\
					\cmidrule(r){2-5} 
					\textbf{MPC Objective} & \textbf{Button Press} & \textbf{Lever Pull} & \textbf{Stack} & \textbf{Square} \\
					\midrule
					Cosine similarity of aligned slots & \textbf{0.64} & \textbf{0.52} & \textbf{0.44} & \textbf{0.22} \\
					SSE of aligned slots & \textbf{0.64} & \textbf{0.52} & \underline{0.42} & \textbf{0.22} \\
					SSE of aligned slot masks & 0.58 & 0.50  & 0.30 & 0.10 \\ 
					\bottomrule
				\end{tabular}
			\end{threeparttable}
		\end{center}
	\end{subtable}
\end{table}